\documentclass[10pt,twocolumn,letterpaper]{article}

\usepackage[pagenumbers]{cvpr} 

\usepackage[dvipsnames]{xcolor}

\definecolor{cvprblue}{rgb}{0.21,0.49,0.74}
\usepackage[pagebackref,breaklinks,colorlinks,citecolor=cvprblue]{hyperref}

\def\paperID{x} 
\def\confName{x}
\def\confYear{x}

\usepackage{makecell}
\usepackage{graphicx}
\usepackage{amsmath, bm}
\usepackage{tabularx,booktabs}
\usepackage{adjustbox}
\usepackage{multirow}
\usepackage{graphicx,xcolor,colortbl} 
\definecolor{LightCyan}{rgb}{0.88,1,1}
\usepackage{array}
\newcolumntype{C}[1]{>{\centering\arraybackslash}p{#1}}
\newcolumntype{L}[1]{>{\arraybackslash}p{#1}}
\usepackage{indentfirst}
\usepackage{pifont}
\usepackage{xcolor}
\usepackage{hyperref}

\title{Amodal Depth Anything: Amodal Depth Estimation in the Wild}

\author{
Zhenyu Li$^1$, Mykola Lavreniuk$^2$, Jian Shi$^1$, Shariq Farooq Bhat$^1$, Peter Wonka$^1$ \\
$^1$KAUST, $^2$Space Research Institute NASU-SSAU \\
\small\url{https://zhyever.github.io/amodaldepthanything/} \\
}

\begin{document}

\twocolumn[{%
\renewcommand\twocolumn[1][]{#1}%
\maketitle
\begin{center}
    \centering
    \captionsetup{type=figure}
    \includegraphics[width=\textwidth]{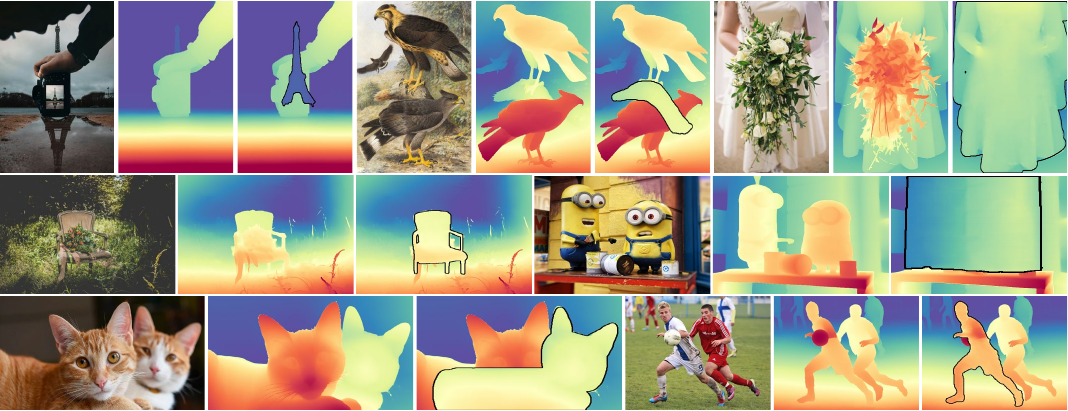}
    \captionof{figure}{\textbf{Amodal Depth Estimation in the Wild.} For each image, we present the general depth estimation result alongside our amodal depth estimation with the target object outlined in black. Our model demonstrates strong generalization across diverse scenes for accurate depth estimation for occluded parts of objects. Best viewed in color.}
    \label{fig:teaser}
\end{center}%
}]

\begin{abstract}

Amodal depth estimation aims to predict the depth of occluded (invisible) parts of objects in a scene. This task addresses the question of whether models can effectively perceive the geometry of occluded regions based on visible cues. Prior methods primarily rely on synthetic datasets and focus on metric depth estimation, limiting their generalization to real-world settings due to domain shifts and scalability challenges. In this paper, we propose a novel formulation of amodal depth estimation in the wild, focusing on relative depth prediction to improve model generalization across diverse natural images. We introduce a new large-scale dataset, Amodal Depth In the Wild (ADIW), created using a scalable pipeline that leverages segmentation datasets and compositing techniques. Depth maps are generated using large pre-trained depth models, and a scale-and-shift alignment strategy is employed to refine and blend depth predictions, ensuring consistency in ground-truth annotations. To tackle the amodal depth task, we present two complementary frameworks: Amodal-DAV2, a deterministic model based on Depth Anything V2, and Amodal-DepthFM, a generative model that integrates conditional flow matching principles. Our proposed frameworks effectively leverage the capabilities of large pre-trained models with minimal modifications to achieve high-quality amodal depth predictions (Fig.~\ref{fig:teaser}). 
Experiments validate our design choices, demonstrating the flexibility of our models in generating diverse, plausible depth structures for occluded regions. Our method achieves a \textbf{69.5\%} improvement in accuracy over the previous SoTA on the ADIW dataset.

\end{abstract}    
\section{Introduction}

Monocular depth estimation is a foundational task in computer vision and generative modeling~\cite{eigen2014mde,bhoi2019monocularsurvey,mertan2022singlesurvey} as it provides depth perception from a single image without stereo cues. 
However, while recent methods, e.g. ~\cite{eigen2014mde,bhat2021adabins,li2022binsformer,yang2024depthanything,bhat2023zoedepth}, focus solely on estimating depth for visible pixels, humans can intuitively perceive the complete 3D geometry of objects, even when only parts of them are visible. Amodal depth estimation is the task of predicting depth values for the occluded (invisible) parts of objects~\cite{sekkat2024amodalsynthdrive,jo2024occlusion}. In this task, we aim to address this under-explored question: \textbf{\textit{Can models effectively perceive the geometry of the invisible parts of objects in a scene?}}

Unlike inpainting methods~\cite{engstler2024invisiblestitch,zhan2020tfill,shih20203dphoto}, which reconstruct missing image regions, amodal depth estimation is a novel task that extends traditional amodal segmentation~\cite{zhan2024amodalseg,zhan2020pcnet,ozguroglu2024pix2gestalt} into the depth domain, by predicting the depth of occluded object parts. In this task, given an input image with a target amodal mask, the objective is to infer depth values for the occluded object regions. While amodal segmentation benefits from human annotations to collect training samples~\cite{sekkat2024amodalsynthdrive,jo2024occlusion}, there is currently no device capable of collecting ground-truth depth data for occluded parts of objects at scale in real-world scenes.

\begin{figure}
    \centering
    \includegraphics[width=0.95\linewidth]{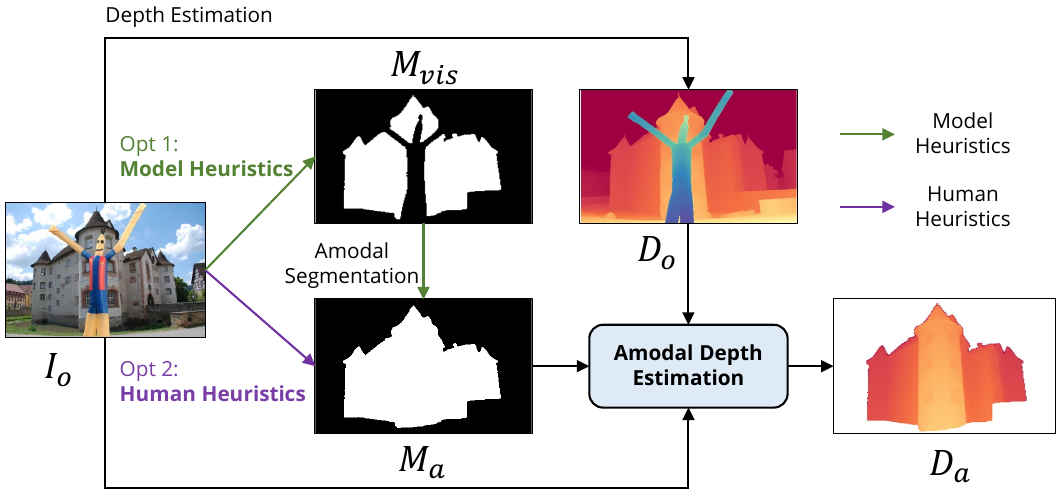}
    \caption{\textbf{Amodal Depth Estimation Pipeline.} Given an input image, users can generate the amodal mask for the depth estimator in two ways: \textbf{\textcolor{PineGreen}{(1) Model Heuristics:}} click the target object, apply SAM~\cite{kirillov2023sam} to generate modal mask, then use amodal segmentation methods to estimate amodal mask, \textbf{\textcolor{Plum}{(2) Human Heuristics:}} manually draw the amodal mask. Our model estimates amodal depth based on original observation image $I_o$, the observed depth map $D_o$, and the amodal mask $M_a$.}
    \label{fig:infer_pipe}
\end{figure}

To address this challenge, prior approaches to amodal depth estimation have relied on synthetic datasets~\cite{sekkat2024amodalsynthdrive,jo2024occlusion}. However, generating these datasets is both time-consuming and hard to scale, often requiring manually placing occluders one by one~\cite{sekkat2024amodalsynthdrive,jo2024occlusion}. Additionally, synthetic data lacks the complexity and diversity of real-world scenes, resulting in a domain gap that limits the generalization of models trained on such data. Furthermore, these previous approaches only consider metric depth, where the goal is to estimate real-world distances for occluded regions. Reliance on metric depth, which naturally struggles to generalize to unseen scenarios with limited data~\cite{piccinelli2024unidepth,bochkovskii2024depthpro}, exacerbates the models’ poor zero-shot performance on real-world images~\cite{yang2024depthanything,Ranftl2022midas,birkl2023midas31}.

To overcome these limitations, we propose a novel formulation of \textbf{amodal depth estimation in the wild}, focusing on \textit{relative} depth. Recent advancements in depth estimation models, such as Depth Anything~\cite{yang2024depthanything}, have enabled the generation of high-quality relative depth maps from natural images. Focusing on relative depth allows us to train models using real-world data, leveraging the depth relationships within scenes to achieve better generalization without relying on precise metric measurements.

We introduce a new large-scale real-world dataset, \textbf{A}modal \textbf{D}epth \textbf{I}n the \textbf{W}ild (\textbf{ADIW}), to facilitate the training of models for amodal depth estimation. Our data generation pipeline leverages segmentation datasets to create high-quality datasets for amodal relative depth estimation with scalable annotations. We adopt a compositing approach to create training pairs, where objects are sampled and composited into scenes. Corresponding depth maps are estimated using the large depth foundation model~\cite{yang2024depthanythingv2}. Given the challenges posed by occlusion in relative depth estimation, we employ a scale-and-shift alignment technique to refine and blend the depth values, ensuring consistency in the final ground-truth depth map used for training.

In this work, we present two complementary frameworks for amodal relative depth estimation, targeting both deterministic and generative model classes. For deterministic models, we propose Amodal-DAV2, an extension of the Depth Anything V2 (DAV2) model~\cite{yang2024depthanythingv2}. This framework incorporates additional guidance layers to reason about depth in occluded regions. For generative models, we present Amodal-DepthFM, an adaptation of the DepthFM model~\cite{gui2024depthfm}, which uses a conditional flow matching approach to estimate depth in occluded regions. In both frameworks, we leverage the power of large pre-trained models with minimal modifications to effectively guide them toward predicting depth for occluded parts of objects.

Experimental results indicate that our method achieves SoTA amodal depth estimation on the ADIW dataset and demonstrates strong zero-shot performance on real-world images (Fig.~\ref{fig:teaser}). Comprehensive ablation studies further validate the effectiveness of our design choices, showing that object-level supervision and guidance signals significantly enhance prediction accuracy. Moreover, we demonstrate the flexibility of our models in generating diverse plausible structures for occluded regions, illustrating their adaptability to different occlusion scenarios. In summary, our contributions are:

\begin{itemize} 
    \item A novel task formulation of amodal depth estimation focusing on relative depth, enabling improved generalization capabilities compared to previous metric-based amodal depth estimation. 
    \item ADIW, a large-scale real-world dataset, generated from real-world images using a scalable pipeline with segmentation datasets and compositing techniques, coupled with a scale-and-shift alignment strategy. 
    \item Two novel frameworks for amodal depth estimation, Amodal-DAV2 and Amodal-DepthFM, leveraging large-scale pre-trained models with minimal modifications to achieve high-quality amodal depth predictions. 
\end{itemize}
\section{Related Work}
\label{sec:related}

\subsection{Monocular Depth Estimation}

Monocular depth estimation, which predicts depth from a single image, has made substantial progress in recent years~\cite{eigen2014mde,bhat2021adabins,bhat2023zoedepth,yang2024depthanything,ke2024repurposing,li2023patchfusion}. Early methods focused on domain-specific metric depth estimation, assuming that training and test images share similar characteristics~\cite{eigen2014mde,fu2018dorn,li2022binsformer,li2023depthformer,bhat2022localbins}. However, this in-domain focus poses challenges for generalization to unseen domains. To address this, recent research has shifted toward cross-domain relative depth estimation, where models infer the relative depth relationships between pixels~\cite{Ranftl2022midas,birkl2023midas31,bhat2023zoedepth,yang2024depthanything,yang2024depthanythingv2}. For instance, MiDaS~\cite{Ranftl2022midas} aggregates multiple datasets and trains the model in a relative depth setting, achieving superior zero-shot performance. ZoeDepth~\cite{bhat2023zoedepth} demonstrates that a strong relative depth model can effectively generalize to metric depth estimation through fine-tuning. Depth Anything~\cite{yang2024depthanything} adopts a semi-supervised strategy that combines large-scale labeled and unlabeled data to further boost the model performance. Generative approaches have also emerged, with methods like Marigold~\cite{ke2024repurposing, zhao2023vpd, lavreniuk2024evp} repurposing denoising UNet models~\cite{rombach2022sd} alongside fixed VAE encoders~\cite{van2017vqvae} for depth estimation~\cite{ke2024repurposing}, and DepthFM integrates flow matching principles into the depth estimation pipeline~\cite{gui2024depthfm}. Despite advancements in monocular depth estimation, a major challenge remains: depth estimation in occluded regions. Existing methods focus only on visible areas, leaving unseen parts unaddressed. Our work addresses this gap by enabling direct depth estimation for invisible object regions.

\subsection{Amodal Perception}

Prior work in amodal perception has primarily focused on tasks such as inpainting~\cite{zhan2020pcnet,ehsani2018segan,ozguroglu2024pix2gestalt}, segmentation~\cite{ke2021deep,liu2023humans,qi2019amodal,zhu2017semantic}, and detection~\cite{hsieh2023tracking,kar2015amodal}. Recently, amodal depth estimation has emerged as a new challenge in amodal perception~\cite{sekkat2024amodalsynthdrive,jo2024occlusion}, which aims to predict the depth of occluded regions of objects. For instance, \cite{sekkat2024amodalsynthdrive} proposed the AmodalSynthDrive dataset for synthetic driving scenes and adopted VIP-DeepLab~\cite{qiao2021deeplab} with multi-headed outputs to estimate the depth at different occlusion levels. Similarly, \cite{jo2024occlusion} introduced the indoor-focused Amodal-3D-FRONT dataset and an iterative approach leveraging amodal masks to estimate depth in occluded areas. These works rely heavily on synthetic datasets due to the difficulty of capturing ground-truth depth behind occlusions in real-world settings. This reliance on synthetic data limits model generalization to real-world scenes, and their focus on metric depth increases the difficulty of accurate depth prediction in the zero-shot setting. In contrast, our approach adopts a relative depth-based solution using large-scale pre-trained models and realistic synthetically generated datasets, enabling strong generalization to diverse real-world images.

\subsection{Depth Inpainting}

Depth inpainting, or depth completion, traditionally focuses on filling in missing depth values for visible regions based on sparse depth inputs~\cite{tang2024bilateral,wang2024improving,yan2024tri,wang2023lrru,zhang2023completionformer}. Some methods aim to inpaint depth around occlusions to create novel 3D views~\cite{shih20203dphoto,jampani2021slide,engstler2024invisiblestitch}, this differs from amodal depth estimation, which specifically targets object-level occlusions. For example, 3D Photography~\cite{shih20203dphoto} uses shared edge guidance to combine RGB and depth inpainting, while SLIDE~\cite{jampani2021slide} introduces a soft-layering strategy to preserve visual details in novel views. Recently, diffusion models have shown strong image generation capabilities~\cite{rombach2022sd}. Invisible Stitch~\cite{engstler2024invisiblestitch} uses a Stable Diffusion model variant~\cite{rombach2022sd} to fill in missing image regions, generating novel views~\cite{podell2023sdxl}, followed by a depth painting network to fill holes in the depth map with guidance from the novel-view image. Unlike these methods, which implicitly or explicitly rely on color priors and focus on visual quality, our approach directly estimates occluded region depth, prioritizing depth accuracy independently of color information or visual coherence. This distinction shifts the focus from achieving realistic view synthesis to predicting occluded depths with high geometric accuracy, an approach that enhances scene understanding for real-world applications.

\section{Method}
\label{sec:method}

\subsection{Task Definition}
\label{subsec:definition}

Amodal depth estimation extends the concept of amodal segmentation by predicting depth information for occluded areas. In traditional amodal segmentation~\cite{zhan2020pcnet,ozguroglu2024pix2gestalt,zhan2024amodalseg}, given an input image and a segmentation mask of the visible portion of an object, the goal is to predict the complete object mask, including the occluded (invisible) part. Similarly, amodal depth estimation aims to estimate the depth values for the occluded regions, given an input observation image $I_o$, a corresponding observation depth map $D_o$ of the input image, and a target amodal segmentation mask $M_a$.

Different from earlier works that formulate amodal depth estimation as a \textit{metric} depth estimation task~\cite{sekkat2024amodalsynthdrive,jo2024occlusion}, we propose a novel formulation where the goal is to predict the \textit{relative} depth of the occluded parts of the object for the input image \textit{in the wild}. Recent models, such as Depth Anything~\cite{yang2024depthanything,yang2024depthanythingv2} enable to generate high-quality relative depth maps from natural images, allowing us to train amodal depth models using real-world data as shown in Sec.~\ref{subsec:dataset}.

\subsection{Dataset Collection}
\label{subsec:dataset}

We present \textbf{A}modal \textbf{D}epth \textbf{I}n the \textbf{W}ild (ADIW), a large-scale dataset specifically designed for predicting occluded object relative depths in real-world scenes. Creating a natural image dataset at scale for amodal depth estimation is challenging due to the lack of annotations for hidden regions. No existing device can capture ground-truth depth data for occluded parts at scale in real-world settings. Previous efforts have generally relied on synthetic datasets~\cite{sekkat2024amodalsynthdrive,jo2024occlusion}, which, although valuable, fall short in capturing the complexity and diversity of real-world scenes. Additionally, these approaches require manually placing occluders, limiting scalability. ADIW overcomes these by generating training data from real-world images.

\begin{figure}
    \centering
    \includegraphics[width=0.99\linewidth]{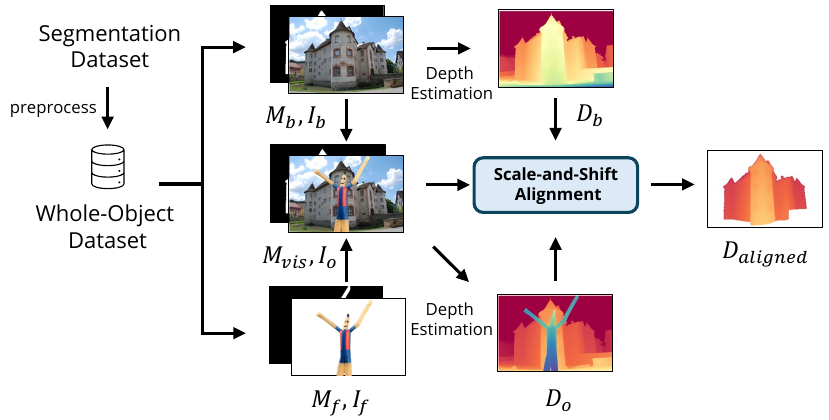}
    \caption{\textbf{Constructing Training Data.} We use the method from \cite{ozguroglu2024pix2gestalt} to convert an initial segmentation dataset into a whole-object dataset. Next, we sample and composite images to create training pairs. Due to occluders, the relative depth predictions differ between the composite and background images, so we apply scale-and-shift alignment for consistent depth blending.}
    \label{fig:create_data}
\end{figure}

We create paired data by overlaying objects over natural images following \cite{ozguroglu2024pix2gestalt}, as illustrated in Fig.~\ref{fig:create_data}. Our method leverages segmentation datasets, which are more scalable than traditional depth estimation datasets~\cite{kirillov2023sam,ravi2024sam2,yang2024depthanything}. Specifically, we employ the Segment Anything model~\cite{kirillov2023sam} to automatically generate segmentation masks from the SA-1B dataset, forming the initial segmentation dataset. Next, we apply the heuristic algorithm from \cite{ozguroglu2024pix2gestalt} to filter out incomplete objects, creating a whole-object dataset.

In our approach, each image $I_b$ (the background image) contains at least one complete object. We then sample an occluder object $I_f$ (the foreground image) and superimpose it onto $I_b$, forming the assembled observation image $I_o$. Both $I_o$ and $I_b$ are then processed through the Depth Anything V2 model~\cite{yang2024depthanythingv2} (ViT-G) to obtain relative depth maps $D_o$ and $D_b$, respectively. Importantly, the depth values for the background object in $D_o$ and $D_b$ differ due to the presence of the foreground object, which changes perceived depth. Both depth maps are scaled to the range $[0, 1]$.

To ensure consistent training labels, we apply the scale-and-shift alignment algorithm~\cite{Ranftl2022midas} to align the background object’s depth values across the two depth maps. The scale factor $s$ and shift factor $t$ are computed as:
\begin{equation}
    (s,t) = \mathrm{argmin}_{s,t} \sum\limits^N_{i=1} (sd_i^b + t -  d_i^o)^2, ~ i \in M_{vis},
\label{eq:ssi-1}
\end{equation}
where $d_i^o$ and $d_i^b$ are the depth values of pixels in the visible part of the background object in $D_o$ and $D_b$, respectively, and $N$ denotes the total number of valid pixels in the visible mask $M_{vis}$ of the background object. The aligned depth map $D_{aligned}$ is then calculated as:
\begin{equation}
    D_{aligned} = sD_b + t,
\label{eq:ssi-2}
\end{equation}
serving as the ground-truth map for model training. This procedure generates a dataset of 564K images with amodal depth labels.

As shown by \cite{yang2024depthanything}, relative depth estimators~\cite{Ranftl2022midas,birkl2023midas31} generalize better than metric depth estimators~\cite{bhat2021adabins,li2022binsformer,li2023depthformer}. Following this insight, we produce relative depth maps for amodal depth estimation in natural scenes. Moreover, our data generation approach is also adaptable for metric amodal depth estimation by omitting the scale-and-shift alignment and utilizing metric depth  models~\cite{piccinelli2024unidepth,bochkovskii2024depthpro}.

\subsection{Amodal Depth Estimator}
\label{subsec:model}

In this work, we aim to leverage large pre-trained depth models by fine-tuning them specifically for amodal depth estimation. Our goal is to make minimal modifications to the original network architectures, preserving the capabilities of the pre-trained weights while introducing the necessary adjustments to guide the model for amodal depth prediction. By integrating additional guidance into the networks, we enable them to predict depth values for occluded parts of objects.

\begin{figure}
    \centering
    \includegraphics[width=0.99\linewidth]{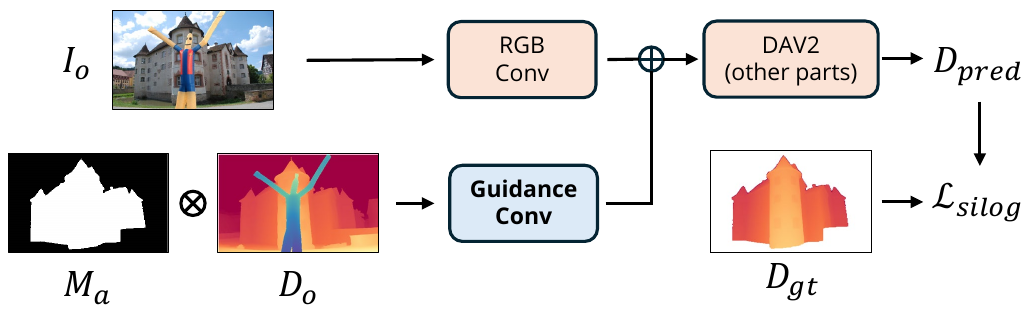}
    \caption{\textbf{Amodal-DAV2 Framework Structure.} Amodal-DAV2 modifies the DAV2 image encoder to take additional guidance channels along with RGB.}
    \label{fig:dav2}
\end{figure}

\begin{figure}
    \centering
    \includegraphics[width=0.99\linewidth]{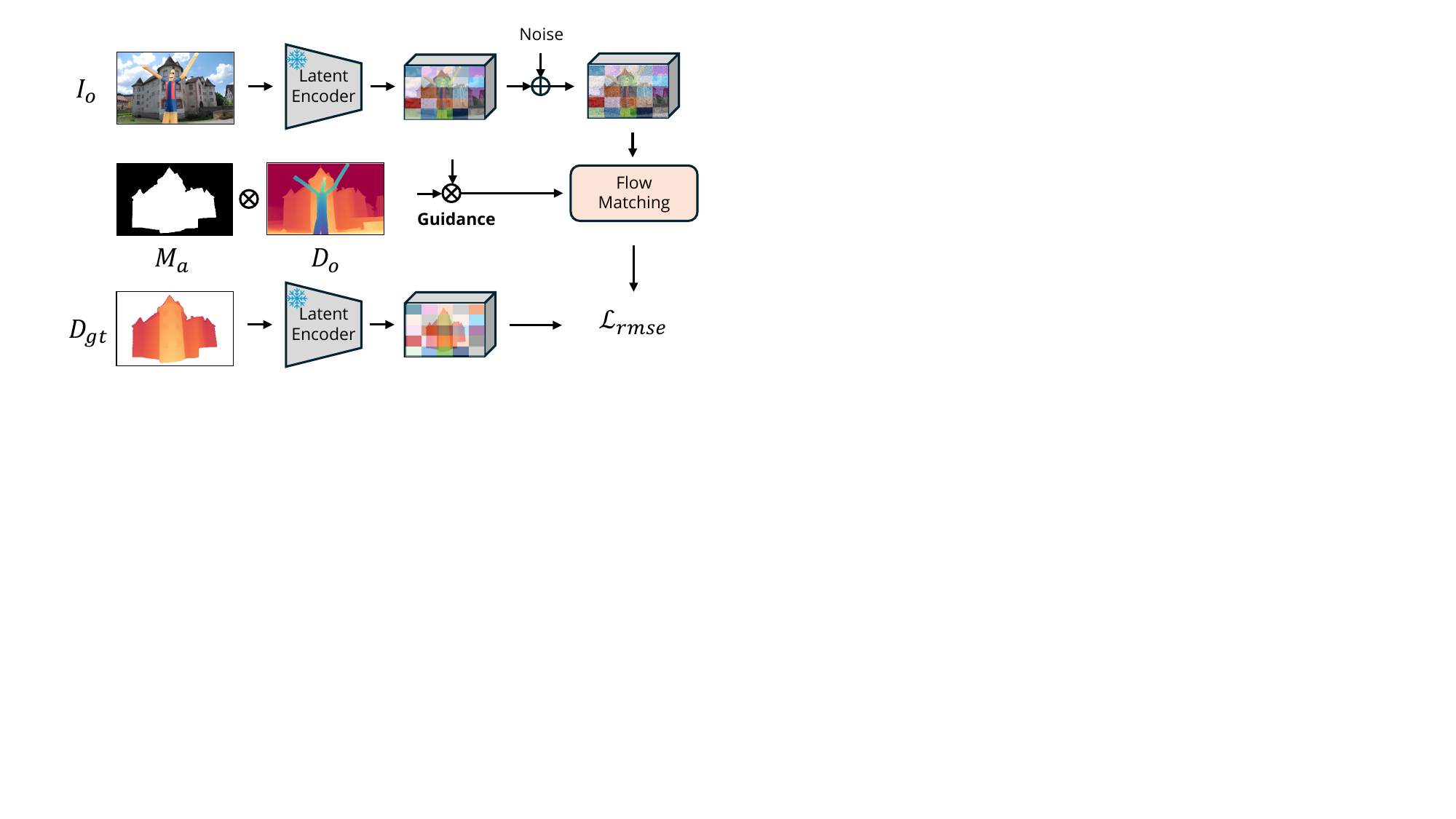}
    \caption{\textbf{Amodal-DepthFM Framework Structure.} Amodal-DepthFM modifies the DepthFM denoising UNet encoder to take additional guidance channels along with RGB latent code.}
    \label{fig:depthfm}
\end{figure}

Depth estimation models fall into two types: deterministic and generative models~\cite{ge2024geobench}. In this section, we introduce a dedicated strategy for adapting each model class to the amodal depth estimation task, as illustrated in Fig.~\ref{fig:dav2} and Fig.~\ref{fig:depthfm}. For the deterministic model, we adopt the Depth Anything V2 (DAV2)~\cite{yang2024depthanythingv2}, a highly representative and top-performing pre-trained model. For the generative model, we select DepthFM~\cite{gui2024depthfm}, known for its superior ability to capture depth details and fast inference speed. 


\subsubsection{Amodal-DAV2}

\noindent \textbf{Model Structure.} The Amodal-DAV2 framework introduces subtle structural modifications to the Depth Anything V2 (DAV2) model's image encoder to retain its pre-trained knowledge while enhancing its capacity in amodal depth estimation. The original DAV2 model employs a Vision Transformer (ViT) architecture~\cite{dosovitskiy2020vit}, which begins with an RGB convolution applied to the input image. As shown in Fig.~\ref{fig:dav2}, we add a Guidance Convolution (Conv)~\cite{sun2024alpha} layer in parallel with the RGB Conv layer, enabling the encoder to accept additional guidance channels for the observation depth map $D_o$ and amodal mask $M_a$. To ensure compatibility with the pre-trained model, the kernel weights of the Guidance Conv layer are initialized to zero, allowing the model to initially ignore the additional guidance information. Additionally, we incorporate layer normalization into the input features of the DPT head~\cite{ranftl2021dpt} to stabilize training and improve convergence.

\noindent \textbf{Training method.} We fine-tune the entire Amodal-DAV2 framework during training, utilizing the standard scale-invariant loss $\mathcal{L}_{si}$~\cite{eigen2014mde} as our objective function:
\begin{equation}
\nonumber
    \mathcal{L}_{si} = \alpha \sqrt{\frac{1}{N}\sum_{i\in M_a} g_{i}^{2} - \frac{\lambda}{N^2}(\sum_{i\in M_a} g_i)^2},
\end{equation}
\noindent where $g_i = \log \Tilde{d_i} - \log d_i$, with $\Tilde{d_i}$ and $d_i$ representing the predicted depth and the ground truth depth, respectively. $N$ denotes the number of valid pixels on the amodal mask $M_{a}$. Although our primary focus is on the accuracy of depth predictions for occluded (invisible) parts of the objects, we supervise the model using the entire object’s depth. This holistic approach helps the model better understand the overall scene structure, leading to improved performance.

\subsubsection{Amodal-DepthFM}
\label{sec:sub:sub:fm}

\noindent \textbf{Framework and Training.} We adapt DepthFM~\cite{gui2024depthfm} for amodal depth estimation using a similar approach. Given pairs of an image $I$, its observation depth map $D_o$, and corresponding amodal mask $M_o$, we fine-tune a conditional flow matching model to achieve amodal depth estimation. The objective is defined as:
\begin{equation}
\nonumber
    \min_\theta\mathbb{E}_{t,z,p(x_0)}||v_\theta(t,\phi_t(x_0))- (x_1-x_0)||,
\end{equation}
where $x_1$ represents the encoded depth samples in the latent space, and the starting point $x_0$ corresponds to an encoded representation of the input image. The latent flow is conditioned on the guidance provided by $z$, which includes the input image latent code, observation depth map $D_o$, and amodal mask $M_a$. Here, $t$ denotes the timestamp, and $\phi_t(x|z)=tx_1+(1-t)x_0$.

During training, we apply noise augmentation, introducing Gaussian noise $\mathcal{N}(x, \sigma_{min})$ to the data samples. Consequently, the conditional probability path is modeled as $p_t(x|z)=\mathcal{N}(x|tx_1+(1-t)x_0,\sigma_{\min}^2\mathbf{I})$. For additional details, we refer readers to \cite{gui2024depthfm}.

In the U-Net architecture used for the flow network $v_\theta$, we modify the first input Conv layer to accommodate the additional guidance channels. Specifically, we extend the channel dimensions to accept extra guidance information $D_o$ and $M_a$ alongside the image latent code. To effectively leverage the pre-trained model, we initialize the first Conv layer with a combination of pre-trained weights for the primary eight channels (corresponding to the depth and image latent codes) and zero-initialized weights for the two additional channels dedicated to guidance inputs $D_o$ and $M_a$.

\noindent \textbf{Scale-and-Shift Alignment Inference.} While the Amodal-DAV2 model directly regresses the amodal depth map, the Amodal-DepthFM model operates in the latent space, predicting the amodal depth latent code, which is then decoded into a depth map using a latent decoder. However, learning consistent amodal depth can be challenging for Amodal-DepthFM. To enhance the predicted depth map, we employ the scale-and-shift alignment~\cite{Ranftl2022midas,rey2022360monodepthtile} during inference.

Rather than relying solely on the model's output, we enhance the prediction by blending it with the observed depth map using a scale-and-shift alignment over the shared visible regions, as described in Eq.~\ref{eq:ssi-1} and Eq.~\ref{eq:ssi-2}. This technique leverages the information from shared regions between the observation depth map and the amodal prediction. By calculating optimal scale and shift factors, we align the depth values of the occluded regions with the visible areas, thereby enhancing the coherence and consistency of the final depth map.

\begin{figure*}[t]
\setlength\tabcolsep{1pt}
\centering
\small
    \begin{tabular}{@{}*{7}{C{2.4cm}}@{}}

    \includegraphics[width=1\linewidth]{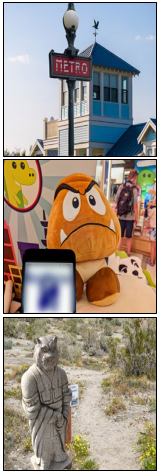} &
    \includegraphics[width=1\linewidth]{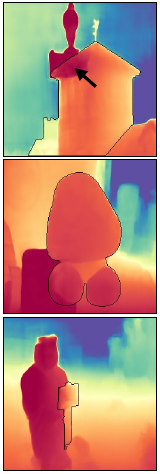} &
    \includegraphics[width=1\linewidth]{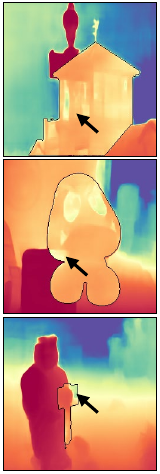} &
    \includegraphics[width=1\linewidth]{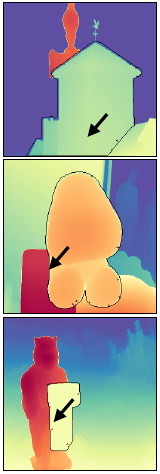} &
    \includegraphics[width=1\linewidth]{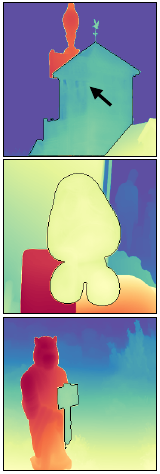} &
    \includegraphics[width=1\linewidth]{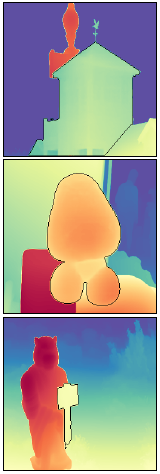} &
    \includegraphics[width=1\linewidth]{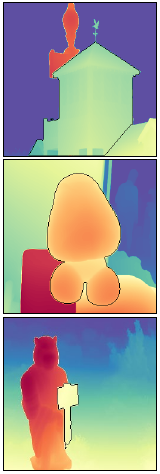} \\
    
    Input  & IS~\cite{engstler2024invisiblestitch} & Jo \etal~\cite{jo2024occlusion} & pix2gestalt~\cite{ozguroglu2024pix2gestalt} & Sekkat \etal$^\dag$~\cite{sekkat2024amodalsynthdrive}  & Ours & GT \\
    \end{tabular}
    \caption{\textbf{Qualitative Comparisons on the Validation Set of ADIW.} Since Invisible Stitch (IS)~\cite{engstler2024invisiblestitch} and Jo \etal~\cite{jo2024occlusion} use metric depth estimation models~\cite{bhat2023zoedepth}, the depth maps are shifted compared with other methods that use relative depth estimators. We evaluate their released models here. $^\dag$: Models retrained with our dataset. Our Amodal-DAV2-L achieves the most consistent scale, accurate shape, and best-preserved details. Best viewed in color for scale accuracy.}
    \label{fig:val}
\end{figure*}

\begin{table*}[t!]
    \centering
    \scalebox{0.90}{
    \begin{tabular}{L{3.1cm}|*{2}{C{1.5cm}}|*{2}{C{1.5cm}}|*{2}{C{1.5cm}}|*{2}{C{1.5cm}}}
        \toprule
        \multirow{2}{*}{Method} & \multicolumn{2}{c|}{Easy} & \multicolumn{2}{c|}{Mid.} & \multicolumn{2}{c|}{Hard} & \multicolumn{2}{c}{Overall} \\
        & RMSE$\downarrow$ & \boldsymbol{$\delta (\%)$}$\uparrow$ & RMSE$\downarrow$ & \boldsymbol{$\delta (\%)$}$\uparrow$ & RMSE$\downarrow$ & \boldsymbol{$\delta (\%)$}$\uparrow$ & RMSE$\downarrow$ & \boldsymbol{$\delta (\%)$}$\uparrow$ \\
        \midrule
        pix2gestalt$^\ddag$~\cite{ozguroglu2024pix2gestalt} & 5.067 & 90.353 & 4.818 & 90.271 & 5.641 & 84.064 & 5.114 & 88.717 \\
        Sekkat \etal$^\dag$~\cite{sekkat2024amodalsynthdrive} & 11.195 & 51.052 & 11.404 & 51.267 & 11.150 & 49.222 & 11.264 & 49.222 \\
        Sekkat \etal$^\dag$$^\ddag$~\cite{sekkat2024amodalsynthdrive} & 5.245 & 88.705 & 5.254 & 88.722 & 5.044 & 87.423 & 5.194 & 88.367 \\
        Jo \etal$^\dag$~\cite{jo2024occlusion} & 10.260 & 56.118 & 10.227 & 55.637 & 9.982 & 54.687 & 10.172 & 55.545 \\
        Jo \etal$^\dag$$^\ddag$~\cite{jo2024occlusion} & 4.624 & 89.848 & 4.777 & 89.915 & 4.728 & 87.256 & 4.712 & 89.177 \\
        \midrule
        Amodal-DAV2-S & 3.574 & 93.650 & 3.741 & 93.666 & 3.734 & 92.125 & 3.682 & 93.251 \\
        Amodal-DAV2-B & 3.482 & 93.871 & 3.620 & 94.002 & 3.640 & 92.622 & 3.578 & 93.590 \\
        Amodal-DAV2-L & \textbf{3.324} & \textbf{94.411} & \textbf{3.460} & \textbf{94.566} & \textbf{3.476} & \textbf{93.309} & \textbf{3.418} & \textbf{94.179} \\
        
        \midrule
        Amodal-DepthFM & 5.354 & 88.427 & 5.398 & 89.241 & 5.498 & 86.970 & 5.410 & 88.353\\
        Amodal-DepthFM$^\ddag$  & 4.622 & 92.494 & 4.500 & 92.982 & 4.883 & 91.048 & 4.645 & 92.295\\
        \bottomrule
    \end{tabular}
    }
    \caption{\textbf{Quantitative Comparisons on ADIW Dataset.} We compare our methods against other possible solutions. Results from Invisible Stitch~\cite{engstler2024invisiblestitch} and Jo \etal~\cite{jo2024occlusion} are not directly comparable, as they use metric depth estimation models~\cite{bhat2023zoedepth}. $^\dag$: Models retrained on our dataset for fair comparison. $^\ddag$: Scale-and-shift alignment applied for consistent prediction handling (see \ref{subsec:model}). Note that Amodal-DAV2 does not rely on this alignment approach. Best results are in \textbf{bold}.}
    \label{tab:overall_table}
\end{table*}

\section{Experiments}
\label{sec:exps}

\subsection{Metrics}

We follow the standard evaluation protocol proposed in previous monocular metric depth estimation works~\cite{eigen2014mde,bhat2021adabins} to evaluate the effectiveness of our framework. We use the root mean squared error (RMSE), the log$_{10}$ error, and the accuracy under the threshold $\delta$. Metrics consider the invisible parts of objects, with the RMSE scaled by a factor of 100 for better illustration. We also categorize objects based on their visible ratios into three difficulty levels: easy $(0.75, 1]$, medium $(0.5, 0.75]$, and hard $(0, 0.5]$. Metrics are calculated separately for each difficulty level to provide a comprehensive evaluation.

\subsection{Implementation Details}

We split our dataset into a training set of approximately 559K samples and a validation set of 4K samples. The Amodal-DAV2 model is trained with a batch size of 32, a learning rate of 1e$^{-5}$, and 50K iterations, using the scale-invariant log (silog) loss with $\lambda$=0.85, following previous works~\cite{bhat2021adabins,li2023depthformer,li2022binsformer}. The Amodal-DepthFM model is trained with a batch size of 128, a learning rate of 3e$^{-5}$, and 15K iterations. Both models are initialized with depth-pretrained parameters before fine-tuning for amodal depth estimation. We use the Adam optimizer with exponential learning rate decay. To stabilize training, we apply a max gradient norm clip of 0.01. Data augmentation is minimal, with only horizontal flipping. All experiments are conducted on 4 NVIDIA A100 GPUs. By default, we evaluate model performance using the final checkpoint after training. 

\subsection{Main Results}

\noindent\textbf{Comparison with Amodal Depth Methods.}
We compare our methods with existing amodal depth estimation approaches. As these methods were initially trained on synthetic datasets and relied on metric depth formulations, we retrained them on our dataset and adapted them to relative depth estimation. Results in Tab.~\ref{tab:overall_table} and Fig.~\ref{fig:val} demonstrate a significant performance gap between the previous SoTA amodal depth model\cite{jo2024occlusion} and our methods, even after output alignment with our strategy. Notably, Amodal-DepthFM outperforms prior approaches even without the alignment, while our best model, Amodal-DAV2-L, improves performance by 27.4\% in terms of RMSE, setting a new SoTA. Previous methods, designed for amodal metric depth estimation, struggle to generalize across varied natural image depth ranges. In contrast, our approach leverages pre-trained DAV2~\cite{yang2024depthanythingv2} and DepthFM~\cite{gui2024depthfm} models, which benefit from extensive prior knowledge of geometry and color, resulting in more accurate amodal depth predictions.

\begin{table}[t!]
    \centering
    \scalebox{0.82}{
    \begin{tabular}{L{0.3cm}|C{2.75cm}|*{1}{C{1.5cm}}|*{1}{C{1.5cm}}|*{1}{C{1.5cm}}}
        \toprule
        &Method & RMSE$\downarrow$ & log$_{10}$$\downarrow$ & \boldsymbol{$\delta (\%)$}$\uparrow$ \\
        \midrule
        \ding{172} & w/o  $D_o, M_{a}$ & 7.549 & 8.607 & 70.607 \\
        \ding{173} & w/o  $M_{a}$ & 4.369 & 4.320 & 91.193 \\
        \ding{174} & w/ align & 3.878 & 3.659 & 92.037 \\
        \ding{175} & Ours (Full) & \textbf{3.682} & \textbf{3.538} & \textbf{93.251} \\
        \midrule
        \ding{176} & $\mathcal{L}_{silog}$ for inv. only & 3.845 & 3.751 & 92.608 \\
        \ding{177} & $\mathcal{L}_{ssi}$ + align & 4.015 & 3.948 & 91.852\\
        \bottomrule
    \end{tabular}
    }
    \caption{\textbf{Ablation Study for Amodal-DAV2.} We investigate the impact of different guidance signals, supervision strategies, and inference techniques.}
    \label{tab:dav2_ablation}
\end{table}

\noindent\textbf{Comparison with Other Solutions.} We also evaluate two alternative solutions: Invisible Stitch~\cite{engstler2024invisiblestitch} and pix2gestalt~\cite{ozguroglu2024pix2gestalt}.
(1) For Invisible Stitch, we use SD-XL~\cite{podell2023sdxl} with a ground-truth amodal mask and image caption to guide the inpainting. The inpainted image serves as guidance for the stitcher model~\cite{engstler2024invisiblestitch} to perform depth inpainting, and the predicted depth is aligned with the observation depth map.
(2) For pix2gestalt~\cite{ozguroglu2024pix2gestalt}, we conduct amodal inpainting followed by depth estimation using DAV2 (ViT-G), aligning the predicted depth with the observation depth. Additional implementation details are in the \textit{supplementary materials}.

\begin{table}[t!]
    \centering
    \scalebox{0.80}{
    \begin{tabular}{L{0.3cm}|C{3.00cm}|*{1}{C{1.5cm}}|*{1}{C{1.5cm}}|*{1}{C{1.5cm}}}
        \toprule
        &Method & RMSE$\downarrow$ & log$_{10}$$\downarrow$ & \boldsymbol{$\delta (\%)$}$\uparrow$ \\
        \midrule
        \ding{172} & w/o align \& $D_o, M_{a}$ & 10.211 & 12.563 & 56.271 \\
        \ding{173} & w/o align \& $M_{a}$ & 5.283 & 6.231 & 83.799 \\
        \ding{174} & w/o align & 5.410 & 5.235 & 88.353 \\
        \ding{175} & Ours (Full) & 4.645 & \textbf{3.553} & \textbf{92.295} \\
        \midrule
        \ding{176} & $\mathcal{L}_{rmse}$ for inv. only & 5.636 & 3.911 & 91.739 \\
        \ding{177} & $\mathcal{L}_{rmse}$ for scene & \textbf{4.608} & 3.809 & 91.149 \\
        \bottomrule
    \end{tabular}
    }
    \caption{\textbf{Ablation Study for Amodal-DepthFM.} We explore the effectiveness of various guidance signals, supervision strategies, and the alignment inference for Amodal-DepthFM. Unlike Amodal-DAV2, the alignment strategy significantly improves the performance of Amodal-DepthFM.}
    \label{tab:depthfm_ablation}
\end{table}

\begin{figure*}[t]
\setlength\tabcolsep{1pt}
\centering
\small
    \begin{tabular}{@{}*{7}{C{2.4cm}}@{}}
    
    \includegraphics[width=1\linewidth]{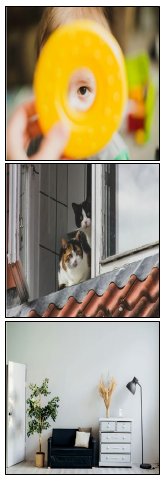} &
    \includegraphics[width=1\linewidth]{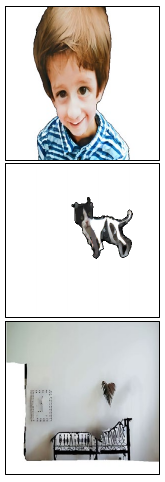} &
    \includegraphics[width=1\linewidth]{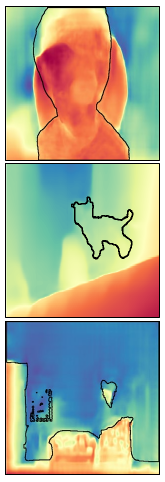} &
    \includegraphics[width=1\linewidth]{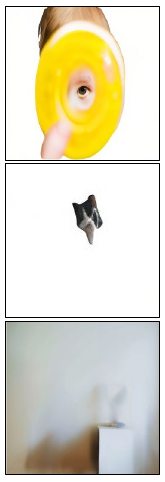} &
    \includegraphics[width=1\linewidth]{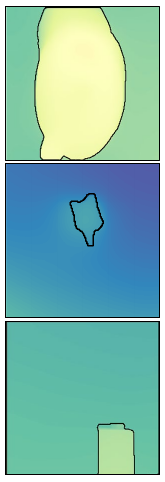} &
    \includegraphics[width=1\linewidth]{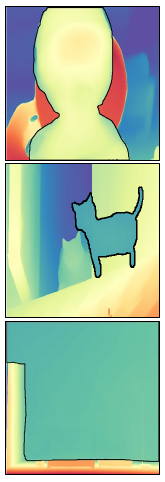} &
    \includegraphics[width=1\linewidth]{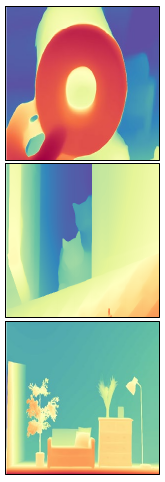} \\
    
    Input  & SD-XL~\cite{podell2023sdxl} & IS~\cite{engstler2024invisiblestitch} & pix2gestalt~\cite{ozguroglu2024pix2gestalt} & pix2gestalt~\cite{ozguroglu2024pix2gestalt} & Ours & $D_o$ \\
    \end{tabular}
    \caption{\textbf{Qualitative Comparisons on Images in the Wild.} Invisible Stitch (IS)~\cite{engstler2024invisiblestitch} uses SD-XL~\cite{podell2023sdxl} with a ground-truth amodal mask and image caption to inpaint occluded areas, while pix2gestalt~\cite{ozguroglu2024pix2gestalt} completes the occluded areas via amodal inpainting. Both methods suffer from inaccurate shapes and cascade depth errors. Our amodal methods estimate depth without any RGB priors~\cite{sekkat2024amodalsynthdrive, jo2024occlusion}, achieving reasonable depth estimations for occluded parts with the guidance of amodal masks.}
    \label{fig:inpainting}
\end{figure*}

\begin{figure}[t]
    \centering
    \includegraphics[width=1\linewidth]{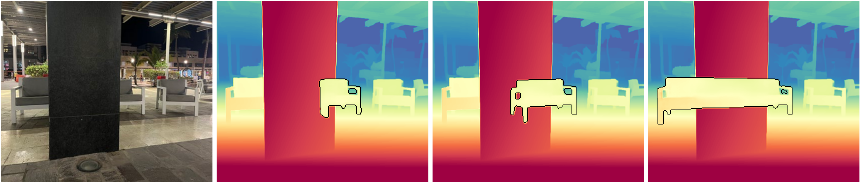}
    \caption{\textbf{Results with Different Mask Guidance.} Guided by various amodal masks, our model successfully predicts the corresponding amodal depth maps for the target objects in the image, showcasing flexibility in estimating occluded regions.}
    \label{fig:diversity}
\end{figure}

As illustrated in Fig.~\ref{fig:val} and Fig.~\ref{fig:inpainting}, the accuracy of amodal depth estimation relies heavily on the quality of RGB inpainting in both methods. Even with the ground-truth amodal mask, SD-XL struggles to accurately inpaint target areas, and the stitcher model fails to predict accurate depth for occluded parts, leaving ghosting artifacts in the predicted depth map. Similarly, pix2gestalt suffers cascade errors due to inaccurate inpainting with uncontrollable amodal shapes. Moreover, since DAV2 is trained on complete natural images, applying it directly to inpainted outputs from pix2gestalt (with single foreground objects on white backgrounds) leads to performance degradation. For example, the window boundary in the first case in Fig.~\ref{fig:val} is missing from the predicted depth map. In contrast, our amodal depth models directly regress the depth of invisible parts without relying on RGB information, using only the amodal mask as guidance. These results demonstrate that our approach provides strong geometric priors, which could also serve as a useful condition for inpainting methods~\cite{zhang2023controlnet}.

\subsection{Ablation Studies and Discussion}
\label{sub:ablation}

\noindent\textbf{Guidance and Supervision Strategies.}
We conduct ablation experiments on Amodal-DAV2 and Amodal-DepthFM to assess each framework component's contribution, with results in Tab.~\ref{tab:dav2_ablation} and Tab.~\ref{tab:depthfm_ablation}, respectively. For both models, guidance from the observation depth map $D_o$ and amodal mask $M_a$ is crucial for optimal performance (\ding{172}, \ding{173}). While the primary goal and evaluation protocol focus is estimating depth for occluded parts of objects, both frameworks benefit from object-level supervision that includes visible regions (\ding{176}, \ding{178}). Interestingly, the scale-and-shift alignment during inference improves Amodal-DepthFM performance but decreases that of Amodal-DAV2, indicating that inconsistent depth estimation may hinder Amodal-DepthFM’s overall performance (\ding{174}).

\begin{figure}[t]
\setlength\tabcolsep{1pt}
\centering
\small
    \begin{tabular}{@{}*{3}{C{2.7cm}}@{}}
    \includegraphics[width=1\linewidth]{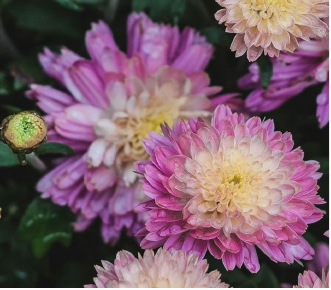} &
    \includegraphics[width=1\linewidth]{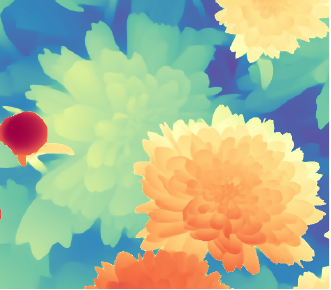} &
    \includegraphics[width=1\linewidth]{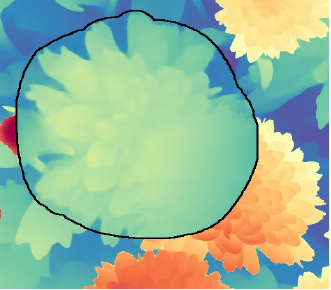} \\
    Input & observation & Amodal-DAV2 \\
    \includegraphics[width=1\linewidth]{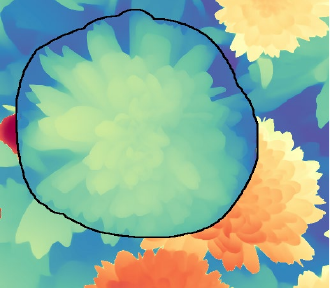} &
    \includegraphics[width=1\linewidth]{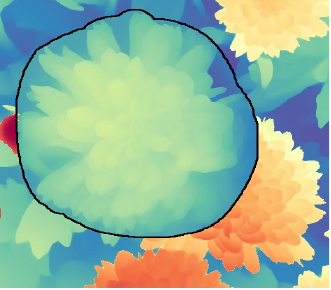} &
    \includegraphics[width=1\linewidth]{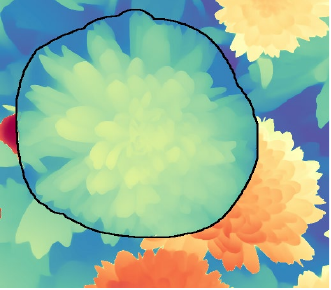} \\
    Amodal-DepthFM 1 & Amodal-DepthFM 2 & Amodal-DepthFM 3 \\
    
    \end{tabular}
    \caption{\textbf{Amodal-DAV2 \textit{vs.} Amodal-DepthFM.} While Amodal-DAV2 achieves better overall performance, Amodal-DepthFM provides more detailed depth with sharper boundaries. Its generative nature allows it to synthesize multiple plausible structures.}
    \label{fig:vs}
\end{figure}

\noindent\textbf{Varying Predictions with Different Guidance.}
Fig.~\ref{fig:diversity} demonstrates the flexibility of our amodal depth estimation model in generating diverse, plausible predictions based on different amodal masks as guidance. This capability is crucial when multiple valid interpretations exist for occluded object parts. For example, the right-hand chair occluded by a wall could vary in width or even be connected to another occluded chair on the left. Our model adapts the amodal depth output according to the provided mask.

\noindent\textbf{Amodal-DAV2 vs. Amodal-DepthFM.}
As shown in Tab.~\ref{tab:overall_table}, Amodal-DAV2 significantly outperforms Amodal-DepthFM, even with scale-and-shift alignment. This indicates that Amodal-DAV2 is generally more effective for amodal depth estimation. However, Fig.~\ref{fig:vs} illustrates that Amodal-DepthFM produces depth maps with finer details and sharper boundaries. Its generative model paradigm allows it to create multiple plausible structures from a given amodal depth map, a capability that Amodal-DAV2, as a deterministic model, lacks.

\section{Conclusion}
\label{sec:conc}

We presented a novel approach to amodal depth estimation, focusing on predicting the depth of invisible parts of objects in natural scenes. Our work introduces Amodal Depth In the Wild (ADIW), a large-scale dataset that leverages segmentation datasets and a compositing pipeline, enabling high-quality, real-world amodal depth annotations. We proposed two complementary frameworks for amodal depth estimation: Amodal-DAV2 and Amodal-DepthFM. Amodal-DAV2 leverages the deterministic capabilities of Depth Anything V2 to achieve state-of-the-art performance in relative depth estimation, while Amodal-DepthFM, built on a generative flow matching paradigm, and excels in providing finer details and sharper boundaries in occluded regions. Our experiments highlight that both models benefit from object-level supervision and the importance of guidance signals for improving amodal depth prediction accuracy. This work not only sets a new benchmark for amodal depth estimation but also opens the door for future research in occluded geometry understanding and improving applications such as inpainting and scene reconstruction.
\appendix

\begin{figure}
    \centering
    \includegraphics[width=0.95\linewidth]{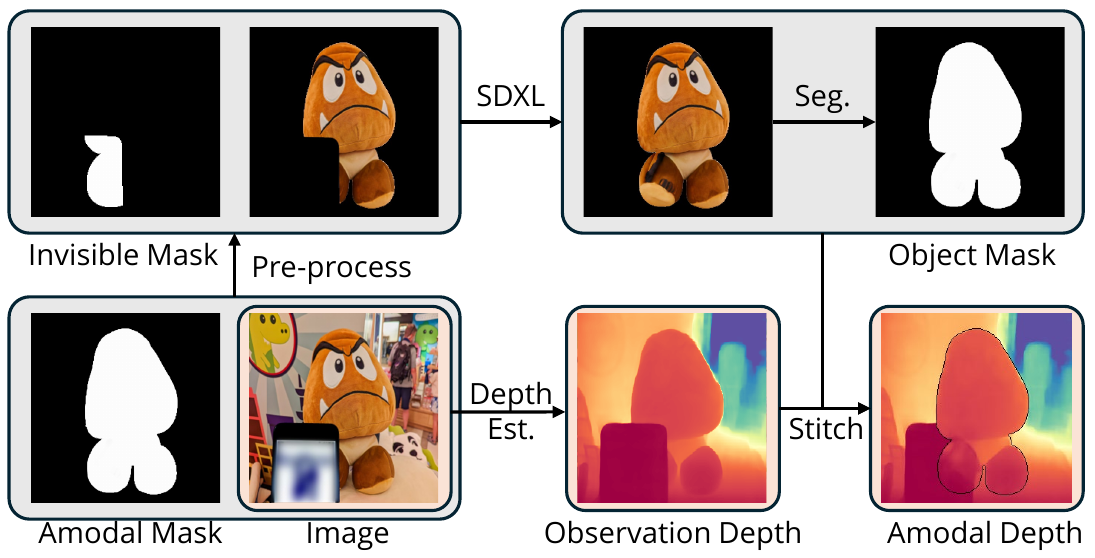}
    \caption{\textbf{Invisible Stitch for Amodal Depth.}}
    \label{fig:infer_pipe1}
\end{figure}

\begin{figure}
    \centering
    \includegraphics[width=0.95\linewidth]{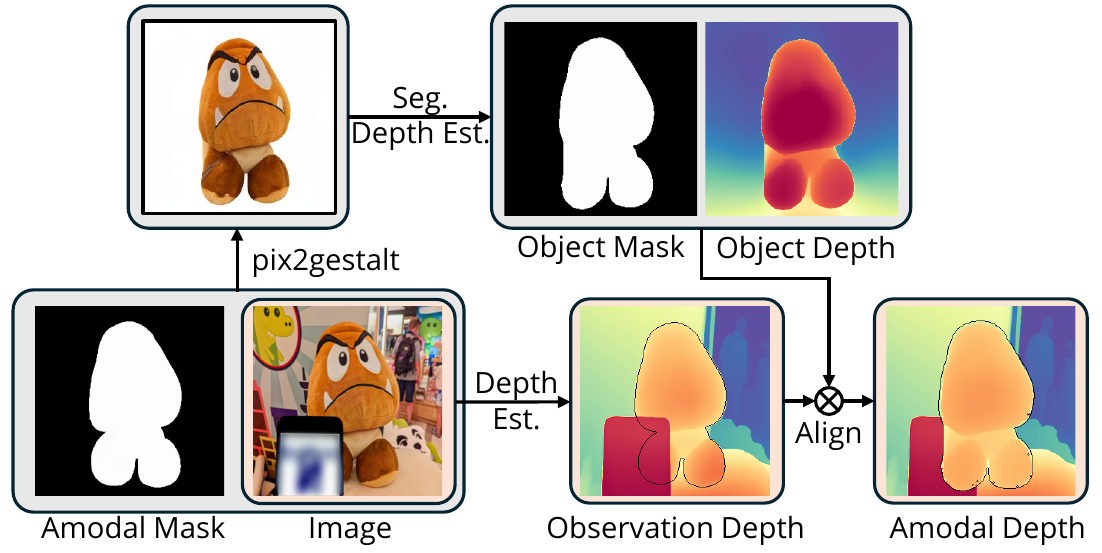}
    \caption{\textbf{Pix2gestalt Stitch for Amodal Depth.}}
    \label{fig:infer_pipe2}
\end{figure}

\begin{figure}[t]
\setlength\tabcolsep{1pt}
\centering
\small
    \begin{tabular}{@{}*{3}{C{2.7cm}}@{}}
    \includegraphics[width=1\linewidth]{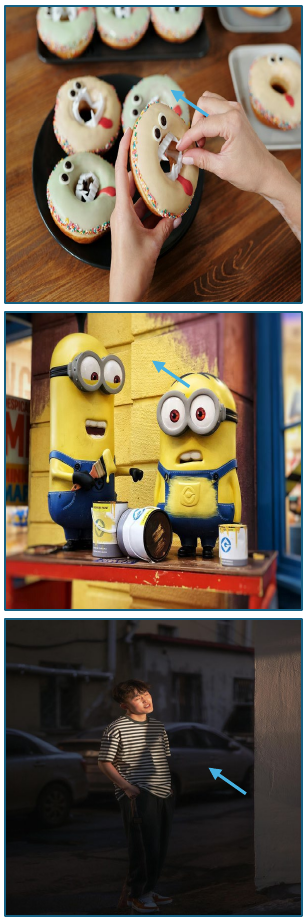} &
    \includegraphics[width=1\linewidth]{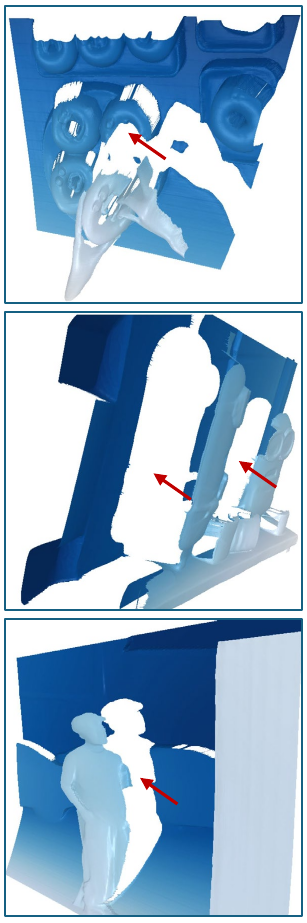} & \includegraphics[width=1\linewidth]{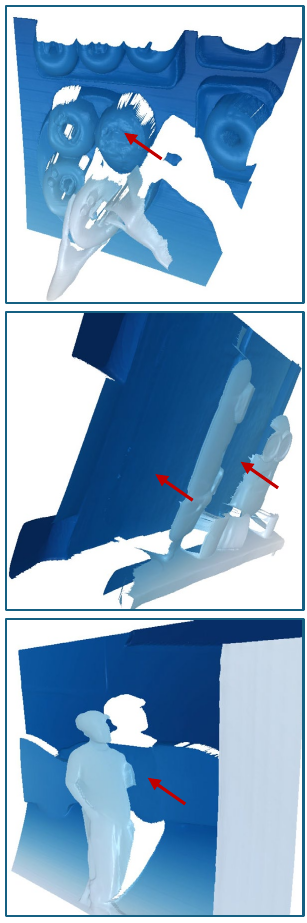} \\
    \end{tabular}
    \caption{\textbf{Reconstructed 3D Mesh for Occluded Object.}  \textcolor{blue}{Blue} arrows indicate the target object and \textcolor{red}{red} arrows highlight the reconstructed meshes for occluded parts of objects, respectively. Left: Input image. Middle: Mesh from general depth. Right: Reconstructed mesh combining general depth and amodal depth.}
    \label{fig:mesh}
\end{figure}

\section{Baseline}
\label{sec:base}

In this section, we describe how two inpainting-based methods, Invisible Stitch~\cite{engstler2024invisiblestitch} and pix2gestalt~\cite{ozguroglu2024pix2gestalt}, are adapted as potential solutions for the amodal depth estimation task.

\subsection{Invisible Stitch}
As illustrated in Fig.~\ref{fig:infer_pipe1}, given an input image and the target amodal mask, we first use SD-XL~\cite{podell2023sdxl} to inpaint the invisible parts of the target object. Note that generating satisfactory inpainting results requires an accurate textual description. For example, we use the textual prompt \textit{“a stuffed toy of an angry-looking character on display”} for the case shown in Fig.~\ref{fig:infer_pipe1}. Next, RMBG-1.4 is applied to create the object mask. Finally, the Invisible Stitch model~\cite{engstler2024invisiblestitch} estimates the observation depth based on the input image and generates the corresponding amodal depth map.

\subsection{pix2gestalt}
The pipeline for adapting pix2gestalt for amodal depth estimation is shown in Fig.~\ref{fig:infer_pipe2}. Given the input image and the target amodal mask, pix2gestalt first inpaints the invisible parts of the target object. Before processing, the amodal mask is preprocessed into the visible mask for the target object. Subsequently,  Depth-Anything V2 ViT-G~\cite{yang2024depthanythingv2} is used to estimate the observation and amodal depth for both the original image and the inpainting result. Finally, the depth maps are aligned using shared visible regions, producing the final amodal depth prediction.

\section{Amodal 3D Reconstruction}
\label{sec:rec}
Using a depth map and the corresponding camera intrinsic parameters, we can reconstruct the 3D point cloud and convert it into a 3D mesh. As shown in Fig.~\ref{fig:mesh}, general depth estimation methods only account for visible pixels, leaving holes in areas corresponding to occluded regions. Our amodal depth estimation provides a simple yet effective solution by predicting reasonable geometry for the invisible parts of objects. This approach can serve as a valuable prior for 3D reconstruction tasks and generative applications, such as novel view synthesis and inpainting.

\section{Limitation and Future Work}
\label{sec:limit}
While our methods demonstrate promising potential for amodal depth estimation, several limitations remain to be addressed in future work. First, our model’s reliance on the input amodal mask, where inaccurate or ambiguous masks can propagate errors and lead to cascading failures in the amodal depth estimation. Additionally, we observe a slight decline in the detail-capturing ability of the depth models after fine-tuning on our dataset. This issue may be attributed to the reliance on the SAM dataset as the foundational dataset. Incorporating diverse, high-quality datasets with more complex target objects could mitigate this limitation and better preserve the model’s ability to capture fine details.

Future directions could extend the single-frame framework to handle videos and develop a unified framework capable of predicting amodal segmentation, RGB, depth, surface normals, and more. These advancements could further enhance the scope and utility of amodal depth estimation.

{
    \small
    \bibliographystyle{ieeenat_fullname}
    \bibliography{main}

\begin{thebibliography}{53}
\providecommand{\natexlab}[1]{#1}
\providecommand{\url}[1]{\texttt{#1}}
\expandafter\ifx\csname urlstyle\endcsname\relax
  \providecommand{\doi}[1]{doi: #1}\else
  \providecommand{\doi}{doi: \begingroup \urlstyle{rm}\Url}\fi

\bibitem[Bhat et~al.(2021)Bhat, Alhashim, and Wonka]{bhat2021adabins}
Shariq~Farooq Bhat, Ibraheem Alhashim, and Peter Wonka.
\newblock Adabins: Depth estimation using adaptive bins.
\newblock In \emph{CVPR}, pages 4009--4018, 2021.

\bibitem[Bhat et~al.(2022)Bhat, Alhashim, and Wonka]{bhat2022localbins}
Shariq~Farooq Bhat, Ibraheem Alhashim, and Peter Wonka.
\newblock Localbins: Improving depth estimation by learning local distributions.
\newblock In \emph{European Conference on Computer Vision}, pages 480--496. Springer, 2022.

\bibitem[Bhat et~al.(2023)Bhat, Birkl, Wofk, Wonka, and M{\"u}ller]{bhat2023zoedepth}
Shariq~Farooq Bhat, Reiner Birkl, Diana Wofk, Peter Wonka, and Matthias M{\"u}ller.
\newblock Zoedepth: Zero-shot transfer by combining relative and metric depth.
\newblock \emph{arXiv preprint arXiv:2302.12288}, 2023.

\bibitem[Bhoi(2019)]{bhoi2019monocularsurvey}
Amlaan Bhoi.
\newblock Monocular depth estimation: A survey.
\newblock \emph{arXiv preprint arXiv:1901.09402}, 2019.

\bibitem[Birkl et~al.(2023)Birkl, Wofk, and M{\"u}ller]{birkl2023midas31}
Reiner Birkl, Diana Wofk, and Matthias M{\"u}ller.
\newblock Midas v3.1 -- a model zoo for robust monocular relative depth estimation.
\newblock \emph{arXiv preprint arXiv:2307.14460}, 2023.

\bibitem[Bochkovskii et~al.(2024)Bochkovskii, Delaunoy, Germain, Santos, Zhou, Richter, and Koltun]{bochkovskii2024depthpro}
Aleksei Bochkovskii, Ama{\"e}l Delaunoy, Hugo Germain, Marcel Santos, Yichao Zhou, Stephan~R Richter, and Vladlen Koltun.
\newblock Depth pro: Sharp monocular metric depth in less than a second.
\newblock \emph{arXiv preprint arXiv:2410.02073}, 2024.

\bibitem[Dosovitskiy et~al.(2020)Dosovitskiy, Beyer, Kolesnikov, Weissenborn, Zhai, Unterthiner, Dehghani, Minderer, Heigold, Gelly, et~al.]{dosovitskiy2020vit}
Alexey Dosovitskiy, Lucas Beyer, Alexander Kolesnikov, Dirk Weissenborn, Xiaohua Zhai, Thomas Unterthiner, Mostafa Dehghani, Matthias Minderer, Georg Heigold, Sylvain Gelly, et~al.
\newblock An image is worth 16x16 words: Transformers for image recognition at scale.
\newblock \emph{arXiv preprint arXiv:2010.11929}, 2020.

\bibitem[Ehsani et~al.(2018)Ehsani, Mottaghi, and Farhadi]{ehsani2018segan}
Kiana Ehsani, Roozbeh Mottaghi, and Ali Farhadi.
\newblock Segan: Segmenting and generating the invisible.
\newblock In \emph{CVPR}, pages 6144--6153, 2018.

\bibitem[Eigen et~al.(2014)Eigen, Puhrsch, and Fergus]{eigen2014mde}
David Eigen, Christian Puhrsch, and Rob Fergus.
\newblock Depth map prediction from a single image using a multi-scale deep network.
\newblock \emph{NeurIPS}, 27, 2014.

\bibitem[Engstler et~al.(2024)Engstler, Vedaldi, Laina, and Rupprecht]{engstler2024invisiblestitch}
Paul Engstler, Andrea Vedaldi, Iro Laina, and Christian Rupprecht.
\newblock Invisible stitch: Generating smooth 3d scenes with depth inpainting.
\newblock \emph{arXiv preprint arXiv:2404.19758}, 2024.

\bibitem[Fu et~al.(2018)Fu, Gong, Wang, Batmanghelich, and Tao]{fu2018dorn}
Huan Fu, Mingming Gong, Chaohui Wang, Kayhan Batmanghelich, and Dacheng Tao.
\newblock Deep ordinal regression network for monocular depth estimation.
\newblock In \emph{CVPR}, pages 2002--2011, 2018.

\bibitem[Ge et~al.(2024)Ge, Xu, Zhao, Sun, Huang, Sun, Chen, and Shen]{ge2024geobench}
Yongtao Ge, Guangkai Xu, Zhiyue Zhao, Libo Sun, Zheng Huang, Yanlong Sun, Hao Chen, and Chunhua Shen.
\newblock Geobench: Benchmarking and analyzing monocular geometry estimation models.
\newblock \emph{arXiv preprint arXiv:2406.12671}, 2024.

\bibitem[Gui et~al.(2024)Gui, Fischer, Prestel, Ma, Kotovenko, Grebenkova, Baumann, Hu, and Ommer]{gui2024depthfm}
Ming Gui, Johannes~S Fischer, Ulrich Prestel, Pingchuan Ma, Dmytro Kotovenko, Olga Grebenkova, Stefan~Andreas Baumann, Vincent~Tao Hu, and Bj{\"o}rn Ommer.
\newblock Depthfm: Fast monocular depth estimation with flow matching.
\newblock \emph{arXiv preprint arXiv:2403.13788}, 2024.

\bibitem[Hsieh et~al.(2023)Hsieh, Khurana, Dave, and Ramanan]{hsieh2023tracking}
Cheng-Yen Hsieh, Tarasha Khurana, Achal Dave, and Deva Ramanan.
\newblock Tracking any object amodally.
\newblock \emph{arXiv preprint arXiv:2312.12433}, 2023.

\bibitem[Jampani et~al.(2021)Jampani, Chang, Sargent, Kar, Tucker, Krainin, Kaeser, Freeman, Salesin, Curless, et~al.]{jampani2021slide}
Varun Jampani, Huiwen Chang, Kyle Sargent, Abhishek Kar, Richard Tucker, Michael Krainin, Dominik Kaeser, William~T Freeman, David Salesin, Brian Curless, et~al.
\newblock Slide: Single image 3d photography with soft layering and depth-aware inpainting.
\newblock In \emph{ICCV}, pages 12518--12527, 2021.

\bibitem[Jo et~al.(2024)Jo, Lee, and Rhee]{jo2024occlusion}
Seong-Uk Jo, Du~Yeol Lee, and Chae~Eun Rhee.
\newblock Occlusion-aware amodal depth estimation for enhancing 3d reconstruction from a single image.
\newblock \emph{IEEE Access}, 2024.

\bibitem[Kar et~al.(2015)Kar, Tulsiani, Carreira, and Malik]{kar2015amodal}
Abhishek Kar, Shubham Tulsiani, Joao Carreira, and Jitendra Malik.
\newblock Amodal completion and size constancy in natural scenes.
\newblock In \emph{ICCV}, pages 127--135, 2015.

\bibitem[Ke et~al.(2024)Ke, Obukhov, Huang, Metzger, Daudt, and Schindler]{ke2024repurposing}
Bingxin Ke, Anton Obukhov, Shengyu Huang, Nando Metzger, Rodrigo~Caye Daudt, and Konrad Schindler.
\newblock Repurposing diffusion-based image generators for monocular depth estimation.
\newblock In \emph{CVPR}, pages 9492--9502, 2024.

\bibitem[Ke et~al.(2021)Ke, Tai, and Tang]{ke2021deep}
Lei Ke, Yu-Wing Tai, and Chi-Keung Tang.
\newblock Deep occlusion-aware instance segmentation with overlapping bilayers.
\newblock In \emph{CVPR}, pages 4019--4028, 2021.

\bibitem[Kirillov et~al.(2023)Kirillov, Mintun, Ravi, Mao, Rolland, Gustafson, Xiao, Whitehead, Berg, Lo, et~al.]{kirillov2023sam}
Alexander Kirillov, Eric Mintun, Nikhila Ravi, Hanzi Mao, Chloe Rolland, Laura Gustafson, Tete Xiao, Spencer Whitehead, Alexander~C Berg, Wan-Yen Lo, et~al.
\newblock Segment anything.
\newblock In \emph{ICCV}, pages 4015--4026, 2023.

\bibitem[Lavreniuk et~al.(2024)Lavreniuk, Bhat, Muller, and Wonka]{lavreniuk2024evp}
Mykola Lavreniuk, Shariq~Farooq Bhat, Matthias Muller, and Peter Wonka.
\newblock Evp: Enhanced visual perception using inverse multi-attentive feature refinement and regularized image-text alignment.
\newblock In \emph{European Conference on Computer Vision Workshops (ECCVW)}, 2024.

\bibitem[Li et~al.(2022)Li, Wang, Liu, and Jiang]{li2022binsformer}
Zhenyu Li, Xuyang Wang, Xianming Liu, and Junjun Jiang.
\newblock Binsformer: Revisiting adaptive bins for monocular depth estimation.
\newblock \emph{arXiv preprint arXiv:2204.00987}, 2022.

\bibitem[Li et~al.(2023{\natexlab{a}})Li, Bhat, and Wonka]{li2023patchfusion}
Zhenyu Li, Shariq~Farooq Bhat, and Peter Wonka.
\newblock Patchfusion: An end-to-end tile-based framework for high-resolution monocular metric depth estimation.
\newblock \emph{arXiv preprint arXiv:2312.02284}, 2023{\natexlab{a}}.

\bibitem[Li et~al.(2023{\natexlab{b}})Li, Chen, Liu, and Jiang]{li2023depthformer}
Zhenyu Li, Zehui Chen, Xianming Liu, and Junjun Jiang.
\newblock Depthformer: Exploiting long-range correlation and local information for accurate monocular depth estimation.
\newblock \emph{Machine Intelligence Research}, pages 1--18, 2023{\natexlab{b}}.

\bibitem[Liu and Vondrick(2023)]{liu2023humans}
Ruoshi Liu and Carl Vondrick.
\newblock Humans as light bulbs: 3d human reconstruction from thermal reflection.
\newblock In \emph{CVPR}, pages 12531--12542, 2023.

\bibitem[Mertan et~al.(2022)Mertan, Duff, and Unal]{mertan2022singlesurvey}
Alican Mertan, Damien~Jade Duff, and Gozde Unal.
\newblock Single image depth estimation: An overview.
\newblock \emph{Digital Signal Processing}, 123:\penalty0 103441, 2022.

\bibitem[Ozguroglu et~al.(2024)Ozguroglu, Liu, Sur{\'\i}s, Chen, Dave, Tokmakov, and Vondrick]{ozguroglu2024pix2gestalt}
Ege Ozguroglu, Ruoshi Liu, D{\'\i}dac Sur{\'\i}s, Dian Chen, Achal Dave, Pavel Tokmakov, and Carl Vondrick.
\newblock pix2gestalt: Amodal segmentation by synthesizing wholes.
\newblock In \emph{CVPR}, pages 3931--3940. IEEE Computer Society, 2024.

\bibitem[Piccinelli et~al.(2024)Piccinelli, Yang, Sakaridis, Segu, Li, Van~Gool, and Yu]{piccinelli2024unidepth}
Luigi Piccinelli, Yung-Hsu Yang, Christos Sakaridis, Mattia Segu, Siyuan Li, Luc Van~Gool, and Fisher Yu.
\newblock Unidepth: Universal monocular metric depth estimation.
\newblock In \emph{CVPR}, pages 10106--10116, 2024.

\bibitem[Podell et~al.(2023)Podell, English, Lacey, Blattmann, Dockhorn, M{\"u}ller, Penna, and Rombach]{podell2023sdxl}
Dustin Podell, Zion English, Kyle Lacey, Andreas Blattmann, Tim Dockhorn, Jonas M{\"u}ller, Joe Penna, and Robin Rombach.
\newblock Sdxl: Improving latent diffusion models for high-resolution image synthesis.
\newblock \emph{arXiv preprint arXiv:2307.01952}, 2023.

\bibitem[Qi et~al.(2019)Qi, Jiang, Liu, Shen, and Jia]{qi2019amodal}
Lu Qi, Li Jiang, Shu Liu, Xiaoyong Shen, and Jiaya Jia.
\newblock Amodal instance segmentation with kins dataset.
\newblock In \emph{CVPR}, pages 3014--3023, 2019.

\bibitem[Qiao et~al.(2021)Qiao, Zhu, Adam, Yuille, and Chen]{qiao2021deeplab}
Siyuan Qiao, Yukun Zhu, Hartwig Adam, Alan Yuille, and Liang-Chieh Chen.
\newblock Vip-deeplab: Learning visual perception with depth-aware video panoptic segmentation.
\newblock In \emph{CVPR}, pages 3997--4008, 2021.

\bibitem[Ranftl et~al.(2021)Ranftl, Bochkovskiy, and Koltun]{ranftl2021dpt}
Ren{\'e} Ranftl, Alexey Bochkovskiy, and Vladlen Koltun.
\newblock Vision transformers for dense prediction.
\newblock In \emph{ICCV}, pages 12179--12188, 2021.

\bibitem[Ranftl et~al.(2022)Ranftl, Lasinger, Hafner, Schindler, and Koltun]{Ranftl2022midas}
Ren\'{e} Ranftl, Katrin Lasinger, David Hafner, Konrad Schindler, and Vladlen Koltun.
\newblock Towards robust monocular depth estimation: Mixing datasets for zero-shot cross-dataset transfer.
\newblock \emph{IEEE TPAMI}, 44\penalty0 (3), 2022.

\bibitem[Ravi et~al.(2024)Ravi, Gabeur, Hu, Hu, Ryali, Ma, Khedr, R{\"a}dle, Rolland, Gustafson, et~al.]{ravi2024sam2}
Nikhila Ravi, Valentin Gabeur, Yuan-Ting Hu, Ronghang Hu, Chaitanya Ryali, Tengyu Ma, Haitham Khedr, Roman R{\"a}dle, Chloe Rolland, Laura Gustafson, et~al.
\newblock Sam 2: Segment anything in images and videos.
\newblock \emph{arXiv preprint arXiv:2408.00714}, 2024.

\bibitem[Rey-Area et~al.(2022)Rey-Area, Yuan, and Richardt]{rey2022360monodepthtile}
Manuel Rey-Area, Mingze Yuan, and Christian Richardt.
\newblock 360monodepth: High-resolution 360deg monocular depth estimation.
\newblock In \emph{CVPR}, pages 3762--3772, 2022.

\bibitem[Rombach et~al.(2022)Rombach, Blattmann, Lorenz, Esser, and Ommer]{rombach2022sd}
Robin Rombach, Andreas Blattmann, Dominik Lorenz, Patrick Esser, and Bj{\"o}rn Ommer.
\newblock High-resolution image synthesis with latent diffusion models.
\newblock In \emph{CVPR}, pages 10684--10695, 2022.

\bibitem[Sekkat et~al.(2024)Sekkat, Mohan, Sawade, Matthes, and Valada]{sekkat2024amodalsynthdrive}
Ahmed~Rida Sekkat, Rohit Mohan, Oliver Sawade, Elmar Matthes, and Abhinav Valada.
\newblock Amodalsynthdrive: A synthetic amodal perception dataset for autonomous driving.
\newblock \emph{IEEE Robotics and Automation Letters}, 2024.

\bibitem[Shih et~al.(2020)Shih, Su, Kopf, and Huang]{shih20203dphoto}
Meng-Li Shih, Shih-Yang Su, Johannes Kopf, and Jia-Bin Huang.
\newblock 3d photography using context-aware layered depth inpainting.
\newblock In \emph{CVPR}, pages 8028--8038, 2020.

\bibitem[Sun et~al.(2024)Sun, Fang, Wu, Zhang, Zang, Kong, Xiong, Lin, and Wang]{sun2024alpha}
Zeyi Sun, Ye Fang, Tong Wu, Pan Zhang, Yuhang Zang, Shu Kong, Yuanjun Xiong, Dahua Lin, and Jiaqi Wang.
\newblock Alpha-clip: A clip model focusing on wherever you want.
\newblock In \emph{Proceedings of the IEEE/CVF Conference on Computer Vision and Pattern Recognition}, pages 13019--13029, 2024.

\bibitem[Tang et~al.(2024)Tang, Tian, An, Li, and Tan]{tang2024bilateral}
Jie Tang, Fei-Peng Tian, Boshi An, Jian Li, and Ping Tan.
\newblock Bilateral propagation network for depth completion.
\newblock In \emph{CVPR}, pages 9763--9772, 2024.

\bibitem[Van Den~Oord et~al.(2017)Van Den~Oord, Vinyals, et~al.]{van2017vqvae}
Aaron Van Den~Oord, Oriol Vinyals, et~al.
\newblock Neural discrete representation learning.
\newblock \emph{NeurIPS}, 30, 2017.

\bibitem[Wang et~al.(2023)Wang, Li, Zhang, Liu, Gao, and Dai]{wang2023lrru}
Yufei Wang, Bo Li, Ge Zhang, Qi Liu, Tao Gao, and Yuchao Dai.
\newblock Lrru: Long-short range recurrent updating networks for depth completion.
\newblock In \emph{CVPR}, pages 9422--9432, 2023.

\bibitem[Wang et~al.(2024)Wang, Zhang, Wang, Li, Liu, Hui, and Dai]{wang2024improving}
Yufei Wang, Ge Zhang, Shaoqian Wang, Bo Li, Qi Liu, Le Hui, and Yuchao Dai.
\newblock Improving depth completion via depth feature upsampling.
\newblock In \emph{CVPR}, pages 21104--21113, 2024.

\bibitem[Yan et~al.(2024)Yan, Lin, Wang, Zheng, Wang, Zhang, Li, and Yang]{yan2024tri}
Zhiqiang Yan, Yuankai Lin, Kun Wang, Yupeng Zheng, Yufei Wang, Zhenyu Zhang, Jun Li, and Jian Yang.
\newblock Tri-perspective view decomposition for geometry-aware depth completion.
\newblock In \emph{CVPR}, pages 4874--4884, 2024.

\bibitem[Yang et~al.(2024{\natexlab{a}})Yang, Kang, Huang, Xu, Feng, and Zhao]{yang2024depthanything}
Lihe Yang, Bingyi Kang, Zilong Huang, Xiaogang Xu, Jiashi Feng, and Hengshuang Zhao.
\newblock Depth anything: Unleashing the power of large-scale unlabeled data.
\newblock \emph{arXiv preprint arXiv:2401.10891}, 2024{\natexlab{a}}.

\bibitem[Yang et~al.(2024{\natexlab{b}})Yang, Kang, Huang, Zhao, Xu, Feng, and Zhao]{yang2024depthanythingv2}
Lihe Yang, Bingyi Kang, Zilong Huang, Zhen Zhao, Xiaogang Xu, Jiashi Feng, and Hengshuang Zhao.
\newblock Depth anything v2.
\newblock \emph{arXiv preprint arXiv:2406.09414}, 2024{\natexlab{b}}.

\bibitem[Zhan et~al.(2024)Zhan, Zheng, Xie, and Zisserman]{zhan2024amodalseg}
Guanqi Zhan, Chuanxia Zheng, Weidi Xie, and Andrew Zisserman.
\newblock Amodal ground truth and completion in the wild.
\newblock In \emph{CVPR}, pages 28003--28013, 2024.

\bibitem[Zhan et~al.(2020{\natexlab{a}})Zhan, Pan, Dai, Liu, Lin, and Loy]{zhan2020pcnet}
Xiaohang Zhan, Xingang Pan, Bo Dai, Ziwei Liu, Dahua Lin, and Chen~Change Loy.
\newblock Self-supervised scene de-occlusion.
\newblock In \emph{CVPR}, pages 3784--3792, 2020{\natexlab{a}}.

\bibitem[Zhan et~al.(2020{\natexlab{b}})Zhan, Pan, Dai, Liu, Lin, and Loy]{zhan2020tfill}
Xiaohang Zhan, Xingang Pan, Bo Dai, Ziwei Liu, Dahua Lin, and Chen~Change Loy.
\newblock Self-supervised scene de-occlusion.
\newblock In \emph{CVPR}, pages 3784--3792, 2020{\natexlab{b}}.

\bibitem[Zhang et~al.(2023{\natexlab{a}})Zhang, Rao, and Agrawala]{zhang2023controlnet}
Lvmin Zhang, Anyi Rao, and Maneesh Agrawala.
\newblock Adding conditional control to text-to-image diffusion models.
\newblock In \emph{ICCV}, pages 3836--3847, 2023{\natexlab{a}}.

\bibitem[Zhang et~al.(2023{\natexlab{b}})Zhang, Guo, Poggi, Zhu, Huang, and Mattoccia]{zhang2023completionformer}
Youmin Zhang, Xianda Guo, Matteo Poggi, Zheng Zhu, Guan Huang, and Stefano Mattoccia.
\newblock Completionformer: Depth completion with convolutions and vision transformers.
\newblock In \emph{CVPR}, pages 18527--18536, 2023{\natexlab{b}}.

\bibitem[Zhao et~al.(2023)Zhao, Rao, Liu, Liu, Zhou, and Lu]{zhao2023vpd}
Wenliang Zhao, Yongming Rao, Zuyan Liu, Benlin Liu, Jie Zhou, and Jiwen Lu.
\newblock Unleashing text-to-image diffusion models for visual perception.
\newblock In \emph{Proceedings of the IEEE/CVF International Conference on Computer Vision (ICCV)}, pages 5729--5739, 2023.

\bibitem[Zhu et~al.(2017)Zhu, Tian, Metaxas, and Doll{\'a}r]{zhu2017semantic}
Yan Zhu, Yuandong Tian, Dimitris Metaxas, and Piotr Doll{\'a}r.
\newblock Semantic amodal segmentation.
\newblock In \emph{CVPR}, pages 1464--1472, 2017.

\end{thebibliography}
}


\end{document}